# Riverine Flood Prediction and Early Warning in Mountainous Regions using Artificial Intelligence

Haleema Bibi[1], Sadia Saleem[1], Zakia Jalil[1]*, Muhammad Nasir[1], Tahani Alsubait[2]

[1]International Islamic University Islamabad, Islamabad, Pakistan

[2]Computer Science and Artificial Intelligence Department, Umm Al-Qura University, Makkah, Saudi Arabia

Corresponding Author: zakia.jalil@iiu.edu.pk,

**Abstract**

Flooding is the most devastating phenomenon occurring globally, particularly in mountainous regions, risk dramatically increases due to complex terrains and extreme climate changes. These situations are damaging livelihoods, agriculture, infrastructure, and human lives. This study uses the Kabul River between Pakistan and Afghanistan as a case study to reflect the complications of flood forecasting in transboundary basins. The challenges in obtaining upstream data impede the efficacy of flood control measures and early warning systems, a common global problem in similar basins. Utilizing satellite-based climatic data, this study applied numerous advanced machine-learning and deep learning models, such as Support Vector Machines (SVM), XGBoost, and Artificial Neural Networks (ANN), Long Short-Term Memory (LSTM) networks, and Gated Recurrent Units (GRU) to predict daily and multi-step river flow. The LSTM network outperformed other models, achieving the highest $R^2$ value of 0.96 and the lowest RMSE value of 140.96 m$^3$/sec. The time series LSTM and GRU network models, utilized for short-term forecasts of up to five days, performed significantly. However, the accuracy declined beyond the fourth day, highlighting the need for longer-term historical datasets for reliable long-term flood predictions. The results of the study are directly aligned with Sustainable Development Goals 6, 11, 13, and 15, facilitating disaster and water management, timely evacuations, improved preparedness, and effective early warning.

**Keywords:** Flood Forecasting, Transboundary River, LSTM, GRU, ANN, Multistep forecasting.

1. Introduction

Floods are one of the most frequent and devastating natural disasters occurring globally, accounting for nearly 40% of all natural disasters (*WHO*, 2024). Floods affect millions of people every year, causing widespread destruction to infrastructure, agriculture, human lives, and ecosystems. Floods contaminate drinking water and spread diseases. In agriculture, floods lead to food shortages and economic losses by ruining livestock and crops. In 2020, affected over 50 million people worldwide, resulting in economic losses worth billions of dollars and significant human suffering (*CRED*, 2024). Third-world nations are the most affected due to the limited resources and poor infrastructure. The global risk of flooding is increasing due to the rising temperatures, which leads to a rise in sea level. In the South Asian region, flooding is a constant concern, and countries like India, Bangladesh, and Pakistan are particularly vulnerable. A total of 9 out of 10 nations with the highest risk of river floods in 2024 were in Southeast Asia, making the area especially susceptible to floods (*Statista*, 2024). Low average elevated positions, recurrent tropical storms, extended monsoons, and insufficient flood protection infrastructure are some of the factors that contribute to this vulnerability. According to the World Risk Index (WRI) 2024 assessment, Pakistan is one of the nations that are most vulnerable to conflict and one of the top 15 nations at risk of disaster, having a risk index score of 9.5 (*DAWN*, 2024); (*WorldRiskReport*, 2024). At the same time, Pakistan is the fifth most susceptible nation to climate change and severe weather (*UN Pakistan Country Report*, 2023), despite its contribution of less than 1% in global carbon emissions (*UN Pakistan Country Report*, 2023). Pakistan, in particular, has suffered recurrent flood disasters, especially in the Khyber Pakhtunkhwa (KPK) region. The terrible 2010 floods have affected over 20 million people, and inflicted an estimated loss of $10 billion. The primary reasons for the 2010 floods were a combination of extreme monsoon rains, rapid glacial melting in the mountains, and unpredictable weather patterns. The flood event also highlighted the vulnerabilities of the water system of the Indus River Basin as well as its tributaries, including the Kabul River (A. Hussain et al., 2022). More recently, in 2022, Pakistan experienced one of its most catastrophic flood seasons, with 33 million people affected, 1,700 deaths, and widespread agricultural and infrastructural damage (*Pakistan: Floods | ReliefWeb*, 2022). In flood-prone areas like KPK and the Kabul River Basin, these regular floods highlight the critical need for efficient flood forecasting and risk management measures. Khyber Pakhtunkhwa, which is located in northwestern Pakistan, is among the areas most vulnerable to flooding due to mountainous terrain, seasonal rainfall, and rivers like the Kabul. The Kabul River runs from Afghanistan into Pakistan and is generally very prone to heavy floods during the monsoon season. This river is a source of water supply for irrigation and agriculture in the region, but its levels are rising uncontrollably posing a great risk (Baig & Hasson, 2023). Considering the effects of climate change getting stronger, the chances of such floods are expected to increase with more frequent and intense precipitation, increased temperatures, and accelerated glacier melting, leading to higher river flows and flood risks(*WorldRiskReport*, 2024). In August 2022, the Swat and Panjkora rivers, which are major sub-basin of the Kabul River, had devastating floods as a result of heavy rain in the Swat and Dir districts, uprooting homes, businesses, hotels, restaurants, stores, and marketplaces. The floodwaters rose to hazardous levels, which had an impact on the water level of the Charsadda and Nowshera districts impacted by the Kabul River. This devastating situation compelled people in a large number to evacuate from their places (Khaliq et al., n.d.), highlighting the necessity of accurate flood forecasting and mitigation strategies to be used in order to counteract the frequent flood effect of such disasters (Islamic Relief, 2022), (Ali et al., 2024). Flood forecasting in Kabul River basin remains a challenge due to political conflicts, less availability of data, and difficult terrain of this region. Lack of proper hydrological modeling in the Kabul River Basin is observed due to limited availability of data for much of the existing work (A. Hussain et al., 2022) (*Pakistan: Floods | ReliefWeb*, 2022). This emphasizes the urgent requirement for more accurate flood risk models and coordinated efforts from Pakistan and Afghanistan. Flood forecasting in the Kabul River Basin has traditionally relied on hydrological models like rainfall-runoff models and physically-based models, which simulate the processes of precipitation accumulation, runoff, and snowmelt to predict river flow and flood risk. A model has been developed using two-dimensional diffusion wave equations to assess flooding and expelled from rain in the Kabul watershed(Sayama et al., 2012). This model reproduced all the flooding processes and produced an understanding of flood dynamics associated with the 2010 Pakistan floods as compared to the satellite-derived inundation maps. A rainfall-runoff-inundation model is utilized to predict the distribution of flood inundation in the Kabul River basin, using corrected rainfall data from rain gauges as inputs to the Global Satellite Mapping of Precipitation (GSMaP) (Ushiyama et al., 2014). The result suggests that the RRI model may successfully predict flood extent, which agrees with Moderate Resolution Imaging Spectroradiometer (MODIS) satellite observations. Another study focused on applying the Integrated Flood Analysis System (IFAS) to anticipate floods, particularly in the Kabul River watershed, after disruptions that occurred in gauge data for the 2010 flood event (Aziz, 2014). The research showed that satellite rainfall data could be used to effectively support rainfall-runoff modeling in minimizing damage from floods. The hydraulic modeling studies related to the lower part of the sub-basins of the Kunar and Kabul have been applied for river flow simulation and assessment of flood risks. The models Hydrologic Engineering Center's River Analysis System (HEC-RAS) in Kabul River basin, Afghanistan, used the different factors that impact flooding, such as land use, soil, and rainfall, to produce flood risk maps and mitigation strategies (Barez, 2023). Although these models have been valuable in understanding the dynamics of floods; however, they have significant limitations. Rainfall-runoff models, for instance, typically assume a linear relationship between rainfall and runoff, which oversimplifies the actual, more complex interactions between climate variables in the Kabul River Basin (Fowler et al., 2020). Non-linear correlations between temperature, snowmelt, and streamflow in hilly areas are sometimes difficult for these models to account for adequately. Additionally, physical models require consistent and trustworthy data for

accurate prediction. However, in remote areas, data collection is difficult due to geopolitical challenges, which hinder the prediction accuracy in the Kabul River Basin (Bazarov et al., 2023). Furthermore, due to the static nature of conventional simulation models, they cannot adapt dynamically to new data and shifting weather conditions. In KPK, where abrupt variations in precipitation or snowmelt may drastically change river flow and flood danger in a short period, this limitation becomes particularly troublesome. Conclusively, these models are less appropriate for forecasting real-time flood or climate change adaptation and planning. To cope with the limitations of such conventional models, this research demonstrates the viability of using time series deep learning and machine learning methods to predict River flow and floods in the Kabul River Basin, Pakistan. Traditional models heavily depend on pre-defined physical relationships between the attributes, while AI-based models can learn directly from data by finding intricate, nonlinear relationships between variables. It has been noticed that deep learning techniques, like ANNs and Recurrent Neural Networks (RNNs), alongside more conventional machine learning techniques like SVMs and Extreme Boosting (XGBoost) are particularly very successful at predicting complex river flow systems. SVMs perform very well compared to traditional hydrological models and are used for regression tasks as they significantly identify the best hyperplane to divide data points in high-dimensional space (Khan et al., 2023). The ANNs perform excellently in terms of learning from large datasets. The models find complex patterns and are very accurate for a wide range of forecasting scenarios (Kasyanto et al., 2023; Rostami & Gholizadeh, 2023). Sequential models such as LSTMs can recall the long-term dependencies in the data. LSTMs can handle sequential data and are good at capturing temporal relationships in data for time series forecasting. The Gated Recurrent Unit (GRU) models are similar to LSTMs but have a simpler design, hence computationally cost-efficient. GRUs capture temporal dependencies effectively while offering equivalent performance with less processing power (Luppichini et al., 2024). These models are capable of dealing with huge data sets, including satellite imagery, real-time weather observations, and historical climate data, making them very versatile to the diverse hydrological conditions of the Kabul River Basin. Additionally, these models are capable of performing real-time forecasting. Hence, flood predictions can be updated continuously based on new data. Furthermore, these models are capable of improving over time as more data is fed into them for training. In contrast with the conventional models, which only learn from the predefined relationships between variables, these models can adapt to the changes in the new data. This characteristic is particularly significant for flood prediction in regions like KPK, which are highly vulnerable to flooding because of gradual changes in rainfall and snowmelt due to climate change and global warming.

In this study, machine learning models, SVM, ANN, XGBoost, LSTM, and GRU have been utilized to predict daily and multistep flood discharge in the Kabul River. Our model affectively predicts the daily river discharge as well as for four days ahead in advance. These early predictions are aligned with SDG 6 (Clean Water and Sanitation), SDG 11 (Sustainable Cities and Communities), SDG 13 (Climate Action) and SDG 15 (Life on Land). Through timely and early forecasts study contributes to effective management of water resources (SDG 6), early warning, and preparedness while reducing the loss of lives and damage to infrastructure (SDG 11). It also contributes to adapting climate change policies (SDG 13), and proper management of water levels to protect the biodiversity and ecosystem along the river.

The rest of the paper is organized in the following way: A brief review of the literature is provided in Section 2. An overview of the study area, dataset, and machine learning and deep learning techniques used in the study is then discussed in Section 3. Section 4 provides the details about the method for implementing the experiment and evaluation matrices. Discussion on results as well as comparison of models is provided in section 5. Section 6 provides the conclusions of the study and future recommendations.

## 2. Literature Review

Traditional statistical and hydrological models using historical patterns and physical characteristics have been utilized for forecasting flood events. The Soil and Water Assessment Tool (SWAT) and the Hydrologic Engineering Center's River Analysis System (HEC-RAS) are used to model the dynamics of River Flow and rainfall-runoff (Anuruddhika et al., 2025; Nguyen et al., 2020). These models rely on meticulous calibration and empirical equations. They offer useful information; however, these models often have shortcomings, including limited real-time adaptability, computational complexity, and dependency on high-quality input data. On the contrary, AI-driven models are excellent at handling nonlinear data relationships. They use huge volumes of meteorological and hydrological data to identify complex data patterns. Hybrid deep learning models, SVMs, and LSTMs have demonstrated significant improvement in early warning and prediction accuracy. Many studies have utilized various AI-driven methods for flood forecasting.

El-Magd et al. investigated machine learning Random Forest, SVM, and ANN models for flash flood forecasting in dry environments (El-Magd et al., 2021). It is found that GIS and remote sensing combined with machine learning can be effectively used to spot flood-prone areas. The findings reveal that Random Forest was the best-performing model for producing accurate forecasts. The LSTM networks and Convolutional Neural Networks (CNN) are combined to evaluate hydrological data for flood prediction (Zhou et al., 2023). This hybrid model produces accurate geographic flood maps with better precision and reduced processing time than conventional techniques. The LSTM is demonstrated as the most significant approach for capturing the underlying patterns in the sequential data (Dtissibe et al., 2024). They used the climatic features, rainfall, and temperature to forecast the floods. For accurate forecasts, it is necessary to gather data from several sources, a large amount of data is required for deep learning models for accurate predictions. For forecasting rainfall, deep learning methods have been investigated from past climate data using models like Deep LSTM (DLSTM) and Deep GRU (DGRU) (Babar et al., 2022). It is proposed that DLSTM outperforms traditional approaches considerably, enabling improved early warning systems in flood control.

By fusing historical climate data, river flow, and machine learning and deep learning techniques, a complete flood forecasting system has been developed (Hayder et al., 2023). Their hybrid exponential-smoothing LSTM strategy speeds up reaction times for early flood alerts, lowers false alarms, and greatly increases forecast accuracy. The SVM, KNN, ANN, and LR are among the data mining approaches evaluated for flash flood prediction, and all of them produced remarkable AUC values over 0.9 (Halim et al., 2022). To improve forecasting applicability, the authors pointed out that domain-specific models that are suited to flash floods are necessary, notwithstanding their efficacy.

Atashi et al. assess how well 1D Convolutional Neural Networks and LSTM networks perform in real-time flood forecasts using historical Red River water level data (Atashi et al., 2023). The findings demonstrate that LSTM performs more accurately than 1D-CNN and successfully captures intricate temporal patterns necessary for prompt emergency responses. For the prediction of flash flood probabilities in North Central Vietnam, 1D-Convolutional Neural Networks with a prediction rate of 90.2% achieved better performance and outperformed conventional models like the SVM and the Logistic Regression(Panahi et al., 2021).

An LSTM-PCA model is introduced that combines Principal Component Analysis and LSTMs for improved flood forecasting (Gunnam et al., 2023). For the examination of the complex meteorological data from the National Climate Data Center, the model achieved an astounding accuracy of 96.49%. An MLP model is utilized for streamflow prediction at the Dayeuhkolot station on the Citarum River for the forecast up to four hours ahead (Dtissibe et al., 2020). It is examined that the regional rainfall data provides more accurate short-term forecasts with R2 and Nash-Sutcliffe Efficiency values higher than 0.9 (Kasyanto et al., 2023).

An LSTM network-based model is designed to forecast flash floods in hilly areas, which performed exceptionally well in predicting peak discharges and has quality ratings of over 82.7% for lead periods of 1 to 10 hours, improving early warning systems (Song et al., 2019). LSTM networks are used to explain Google's operational flood forecasting system for real-time alerts in riverine floods by combining machine learning with real-time data (Nevo et al., 2022). During the 2021 monsoon season, this system successfully notified over 100 million people about the flood. To map flood risks and to evaluate their effectiveness against previous flood events, Random Forest and XGBoost were used. These techniques produce more precise flood susceptibility predictions in flood-susceptible areas(Gharakhanlou & Perez, 2023).

In a novel forecasting method, LSTM is integrated with numerical models for the danger of urban floods (Chen et al., 2023). The proposed model's excellent forecasting accuracy and speedy computations surpass the slow computation speed of traditional models and enable it to efficiently manage floods and react to catastrophes. Dimension reduction technique principal component analysis, combined with a one-dimensional CNN, is used for flood prediction in Kerala, India, using daily rainfall data (John & Nagaraj, 2023). The model achieved an accuracy of 94.24% for flood forecasting while outperforming the previous ones.

For flood forecasting in the Red River region, sophisticated AI algorithms were combined with conventional statistical techniques (Atashi et al., 2022). The findings reveal that the LSTM networks performed best compared to the SARIMA and the Random Forest models, with the lowest RMSE values demonstrating high predictive accuracy over a range of forecasting horizons.

In the Han River in South Korea, a time series model of Gated Recurrent Units is examined to forecast water level (Park et al., 2022). The hydrological data is integrated with meteorological data, which results in the GRU model outperforming LSTM in terms of R2 values ranging from 0.7480 to 0.8318 showing the potential to improve flood prediction through multivariate data.

In Pakistan, the enhancement of flood forecasting through machine learning remains the focus of recent research. Recently, in the Hunza River region, machine-learning algorithms Adaptive Boosting, Random Forest, Gradient Boosting, and K-Nearest Neighbors, were used to predict the monthly stream flow at Danyor Station (M. W. Yaseen et al., 2022). Meanwhile, the adaptive boosting model achieved remarkable performance with an R2 of 0.998. An approach to forecasting floods in the Basin of Chenab River using LSTM and ML-GMDH models is designed (Aatif et al., 2024). The outcomes of the LSTM model show that it is effective in predicting river flow, even in basins with limited gauges. An ensemble (LR-SVM-MLP) model is designed to map Karachi's susceptibility to floods (A. Yaseen et al., 2022). This ensemble model achieved accuracy rates of 98% for testing data and 99% for training. This model significantly enhanced the Water resources and flood control planning. The efficacy of adaptive boosting is demonstrated in forecasting monthly river flow in the basin of the Hunza River (D. Hussain & Khan, 2020). It is examined that AI-based methods need to be improved to strengthen the accuracy of flood forecasts. The ANN and SVM were used to estimate daily discharge and flood occurrences in the Panjkora River sub-basin (Ali et al., 2024). ANN outperformed SVM with an R2 value of 0.75 for daily discharge prediction. Therefore, it is examined that state-of-the-art machine learning models are integrated into the flood forecasting system for improved predictions and disaster management in Pakistan.

In Pakistan's Kabul River basin, recent research has employed various artificial intelligence technologies for flood forecasting, thereby increasing prediction accuracy and enhancing water resource management. A machine learning ARMA model forecasts water levels in the Kabul River for the years 2011–2030 (Musarat et al., 2021) using univariate water data. This data shows seasonal fluctuations at a consistent level of 250 cumecs. However, since the model can only use the historical water data it has been provided, it cannot account for the potential implications of future climate change. To map flood threats in the Peshawar Vale, the authors combine GIS with ANNs, taking into account several flood-causing factors (Saeed et al., 2021). The model demonstrates that an ANN model may accurately depict flood hazard zones; however, it requires verification and a basic understanding of GIS to properly interpret the findings. The LAES model combines ANN and SVM with LASSO feature selection to enhance the forecasting accuracy of the Kabul River's discharge (Shabbir et al., 2023). It outperforms traditional models with an R² of 0.789, but it only incorporates data from Pakistan, despite the majority of the water originating from Afghanistan. The Random Forest Regressor model estimates the daily regime of river flow based on historical climate scenarios and future projections (Din, 2024). The results showed a significant calibration R² of 0.94, indicating that substantial changes in hydrological river flows are anticipated, including increased flooding in the future scenario.

An analysis of previous research on flood forecasting in the Kabul River watershed identifies significant information gaps that must be addressed to accurately and consistently anticipate flooding. The significance and impact of the Afghan portion of the basin on hydrology have not been adequately covered in any research, which has relied solely on data sources from Pakistan. This shortcoming reduces understanding of the overall dynamics of the Kabul River system. Although informative, future water levels were predicted using a single variable and data from the Swat. However, this approach lacked a multivariable perspective, omitting crucial hydrological and climatic factors that influence flood behavior. Conversely, studies that employed both hydrological and meteorological data limited themselves to data from the Pakistan side. Incorporating climatic data from both sides of the transboundary river is essential for understanding the dynamics of Kabul River flow. To date, no study has predicted floods or streamflow using climatic data from both Pakistan and Afghanistan. In this context, time series models have shown superiority in pattern identification and flood prediction in several studies employing both climate and hydrological data. The integration of advanced artificial intelligence techniques could significantly enhance flood prediction accuracy for the Kabul River basin, addressing unique challenges related to its transboundary characteristics and complex hydrological patterns.

### 3. Methodology

The proposed methodology for forecasting daily and multi-day river discharge for flood prediction is explained in this section. The proposed method explains all the steps that contribute to the timely prediction of floods. It involves a discussion of the area selected for the study and how data is collected from that region. Additionally, the preparation of data and the methods used for forecasting are discussed.

### 3.1. Aim and Objective

In mountainous regions, flood prediction poses a significant challenge due to the complex topography, limited data availability, and rapidly changing weather due to global warming. The steep slopes in these areas contribute to faster runoff, which lead to flash floods. The scarcity of hydro-meteorological data and complex infrastructure hindes the flood prediction. Additionally, extreme weather events, such as cloudbursts, glacial melting, and unprecedented rainfall, exacerbate the difficulties in forecasting floods. In transboundary regions, the lack of cooperation regarding data sharing also hampers flood prediction efforts. Addressing this problem is crucial to safeguard lives, infrastructure, and humanity as a whole. This study aims to design a machine-learning-based flood forecasting system for the Kabul River Basin, utilizing advanced algorithms to enhance the accuracy and timeliness of flood predictions. The major objectives of the study are:

1. To explore advanced machine learning techniques for accurately forecasting river discharge to anticipate floods in the mountainous region while using satellite data to overcome data limitations in data data-scarce transboundary basin.
2. To forecast daily river discharge to support real-time flood monitoring multiple days ahead to improve early warning and preparedness.

### 3.2. Study area

The Kabul River Basin, spanning approximately 92,650 km² across eastern Afghanistan and northwestern Pakistan (Iqbal et al., 2018), lies between longitudes 67 ˚ E to 72 ˚ E and latitudes 33 ˚ N to 37 ˚ N, and is a critical hydrological region facing complex water-related challenges. The basin contributes about 16% of Pakistan's total annual water availability (Baig & Hasson, 2023). It is characterized by significant elevation changes, from as low as 249 meters to as high as 7,701 meters above sea level. The region consists of steep mountainous terrain in the north and broad floodplains in the south. This transboundary river basin is home to around nine million people, dependent on its water resources for irrigation, hydropower, and domestic use (SAIS, 2017). The basin's climate is influenced by its diverse topography and converging climatic zones, with precipitation varying from 50 mm in the east to over 600 mm in the west, and mean annual precipitation ranging between 500 mm and 600 mm. Temperatures vary from -10°C in the northern sub-basins to 35°C in the downstream areas (Baig & Hasson, 2023). The Kabul River, stretching over a length of 700 km, flows through Afghanistan. However, the final 140 km crosses the Pakistan border and continues into the Indus River. Its flow is primarily derived from snowmelt and glacial origins, peaking during the spring and summer months, especially in July. Flooding is a significant issue, mostly affecting the low-lying areas downstream. The study area's map is shown in Figure 1, which indicates the locations of stations from which meteorological and hydrological data are collected. The downstream Nowshera station is marked with a red point and is used for discharge data collection.

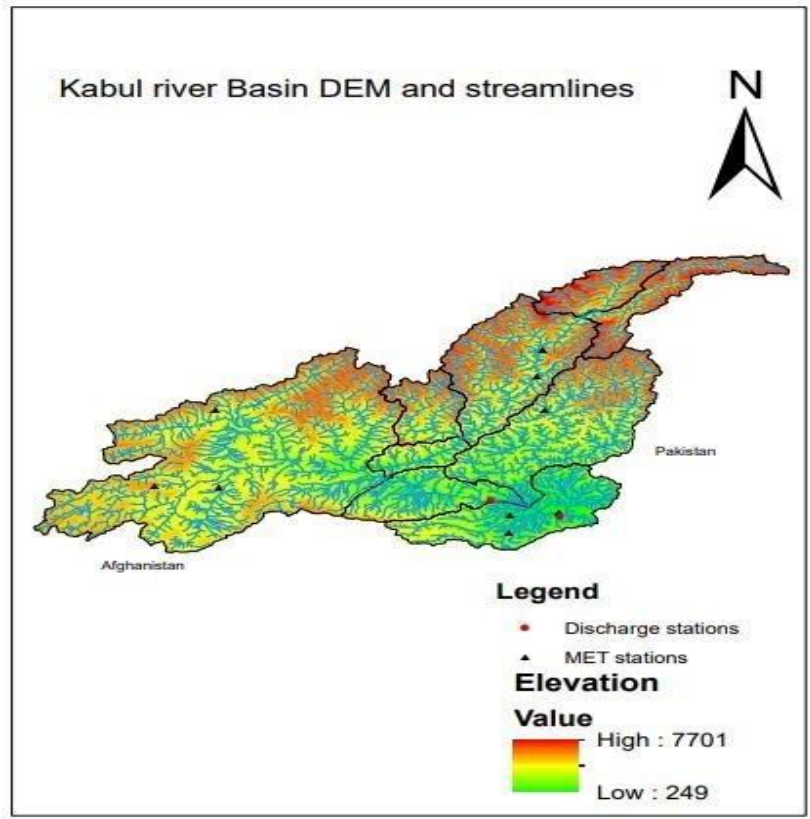

*Figure 1: Map of Kabul River Area*

The 2010 flood was exceptionally catastrophic, resulting in more than 2,000 fatalities and considerable infrastructural destruction (NASA, 2011). Again in 2022 heavy unprecedented rainfall in sub-basins of the Kabul River triggered flooding in the basin which severely affected the Kabul River resultantly most of the area near the river was affected. Simultaneously, it contributes to flood in the major water system Indus River. This catastrophic flood hit more than 33 million people all over the country(*2022 Pakistan Floods*, 2023), therefore, it is need of an hour to design an accurate early warning system and accurate water management in the basin to prevent the losses.

### 3.3. Data Collection

Accurate hydro-meteorological data with high precision is crucial for effective flood forecasting. This remains a persistent challenge in transboundary river basins due to inadequate data sharing between countries(Aatif et al., 2024). This study collected meteorological and hydrological data, including river flow, rainfall, temperature, relative humidity, absolute humidity, and severe weather events in the selected region. This study utilize dataset collected from two primary sources. To obtain meteorological data, the MERRA-2 is used. The data is collected using an average resolution of 0.5 x 0.625 degrees concentrating on specific locations(NASA, GMAO, 2024). The dataset contains numerous climatic variables (temperature, humidity, precipitation, etc.), which are significant to understanding the relationship between climate variables and Water data(Ali et al., 2024). The data spans from the year 2005 to 2020 over multiple areas of Afghanistan and Pakistan. Analysis of data collected utilizing Synchronous Meteorological Satellite (SMS) technology for climate data retrieval helps to gain important insights into the structure of the data(Gelaro et al., 2017). River flow data by the SWHP and the Water and Power Development Authority (WAPDA), including river discharge data along with the date, is accessed. River flow is a crucial target variable for flood prediction. It is measured in cubic meters per second (m³/s), and provides daily records of the Kabul River's flow from 2005 to 2020 at the downstream Nowshera gauging station, which is located at 34 ° 0' 25" to 71 ° 58' 50". Hydrological data is integrated with the meteorological data while maintaining data integrity. The data originated from disparate sources so it was essential to aggregate data from many locations on both sides of the basin to ensure consistency and integrity. The data points were aligned to match the same time and date attributed to the synchronization of the meteorological and hydrological databases. A thorough cleaning procedure is performed on the data to get rid of any inconsistencies, outliers, and duplicates. Following guidance from a WAPDA official representative, special characters in the discharge data were identified and removed. The flow of the proposed methodology is shown in Figure 2.

.

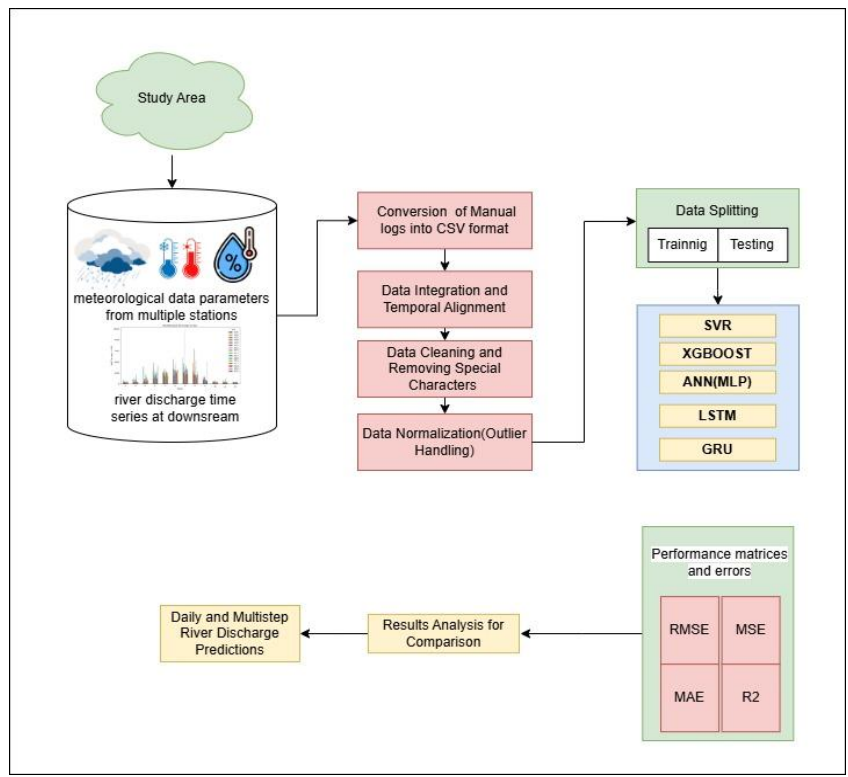

*Figure 2: Architectural Diagram of Early Flood Prediction System for KPK*

## 3.4. Data normalization

The most crucial preprocessing step in machine learning, particularly for time series, is data normalization, which ensures that each attribute makes an equal contribution to the model's learning process. Because it promotes convergence during training and boosts model performance, this phase is crucial. Normalization reduces scale disparities between characteristics by transforming data into a predetermined range or distribution.

- **Robust Scaling**

A statistical method called robust scaling was created especially to pre-process numerical data, especially when the dataset includes outliers. When compared to more conventional techniques like standardization or min-max scaling, it offers a more dependable and robust approach. Robust scaling's fundamental concept is to center the data on the interquartile range, or the middle 50% of the data records. Because the Interquartile Range (IQR) serves as a benchmark, robust scaling is less vulnerable to outliers.

The following are the steps to execute robust scaling:

- Calculate Quartiles: The data set is used to calculate Q1 and Q3. The data is divided into four equal sections by the quartiles.

- Determine the Interquartile Range using Equation (1).

$$IQR: Q3 - Q1 \qquad (1)$$

- Data Scaling: Apply the following formula given in Equation (2) to each data point x:

$$X_{scaled} = \frac{(x - Q1)}{IQR} \qquad (2)$$

The data points are scaled to a range of around -1 to 1 by this scaling method to reduce the influence of outliers. This technique enhances the reliability and stability of the models by preventing extreme values from affecting the models during training.

### 3.5. Data Splitting

The dataset is split into a training and testing set for this time series forecasting task. The ratio of the training data set is 80% of the data, which spans from January 2005 until July 2017. Testing and validation were conducted using the remaining 20% of the data, which spans from July 2017 to December 2020. This sequential data split would allow the model to learn patterns from past data and evaluate it on test data, simulating the real-world scenario of time series forecasting.

### 3.6. Methods

To address the challenges in discharge forecasting in Kabul River, capturing non-linear relationships, and temporal dependencies, and overcoming the impact of possible outliers in the data, different algorithms, including SVR, MLP, XGBoost, LSTMs, and GRUs.

- **Support Vector Regressor (SVR)**

SVR is a renowned technique used for time series regression and forecasting tasks. It finds the best hyperplane that fits the information well while permitting variance. It maps the data into a higher-dimensional space to identify the non-linear correlations between variables significantly by using kernel functions. SVR works well for hydrological forecasting as it can deal with noisy and non-stationary data. By adjusting hyperparameters (kernel function and regularization parameters) to effectively identify complicated irregular patterns in data.

$$Q_t = \sum(\alpha * K[P, Q_{lag}, T_{min}, T_{max}, H], X_i) + b \qquad (3)$$

Where:

- $Q_t$: Predicted discharge
- $P=[P_t, P_{t-1}, P_{t-2}...]$: A vector of precipitation values for the current day and preceding days used as input features.
- $Q_{lag}=[Q_{t-1}, Q_{t-2},...]$: A vector of discharge values from the previous days included as lagged features.
- $[T_{min}, T_{max}, H]$: The current day's minimum and maximum temperature and humidity.
- $K()$: The kernel function.
- $X_i$: Training data-derived support vectors, containing feature sets that include lagged precipitation, lagged discharge, temperature, and humidity values.
- α: The Lagrange multipliers defining the contribution of each support vector.
- b: The bias term learned during training.

The temporal dependencies in the dataset are reflected in the SVR equation. The RBF kernel is used to well capture the nonlinear interactions between variables well, allowing for precise simulation of complex hydrological processes.

- **Multi-Layer Perceptron (MLP)**

MLP is a potential neural network architecture that is inspired by the composition and functions of human brain. In non-linear models, multiple linked layers of neurons are utilized, each of which implements an activation function and the weighted sum of the inputs. MLP may be utilized for hydrological forecasting as it can manage intricate data patterns where nonlinear interactions between the variables may exist. Depending on the task at hand, the number of hidden layers, and neurons in each layer, MLPs may modify their activation functions.

An MLP can be represented mathematically as given below in Equation (4).

$$Y = f(W_1 * W_{\{L-1\}} *...* f(W_L * X + b_1) +...+ b_{\{L-1\}}) + b_l \quad (4)$$

Y: Output of the network

$W_i$: the weight matrix for the $i_{th}$ layer

$b_i$: the bias vector for the $i_{th}$ layer

f: Activation functions (ReLU, sigmoid, tanh)

X: Input vector

The MLP learns to properly anticipate complicated functions by repeatedly adjusting the weights and biases.

- **Extreme Gradient Boosting (XGBoost)**

In recent years, an ensemble learning technique XGBoost has gained a lot of popularity. It is an expansion of the gradient-boosting idea, which builds a collection of decision trees repeatedly. Each tree in the ensemble has been taught to correct the errors of its predecessors to provide a more accurate and dependable model. It uses L1 and L2 regularization methods to prevent overfitting. It uses row sampling, column sampling, and early stopping to further reduce overfitting and increase efficiency.

The following is the definition of the XGBoost objective function:

$$Objective = \sum \left(L(y_i, y'_i)\right) + \sum \Omega(f_t) \quad (5)$$

In given Equation (5)

$L(y_i, y'_i)$: Loss function (Squared error, log loss)

$\Omega(f_t)$: Regularization to penalize model complexity

$y_i$: Value for the ith observation in actual

$\hat{y}_i$: Anticipated value for the $i^{th}$ observation

$f_t$: $t^{th}$ tree in the ensemble

XGBoost is a state-of-the-art method for several problems, including time series tasks, especially well performed when variables have complicated and nonlinear connections.

- **Long Short Term Memory (LSTM)**

LSTM networks are sequential models, certain type of recurrent neural networks. They are very useful for time series forecasting. As compared to RNNs, they do not have the vanishing gradient issue. They are equipped to determine long-term data dependencies. It leverages three gates: the input gate, the forget gate, and the output gate as well as a cell state, which allows information to be preserved for long periods. This model controls amount of data from cell state is output through the output gate and decides which new information to add to cell state and which to forget, i.e. which one will be rejected with the aid of the forget gate. These gates enable LSTMs to model complex temporal patterns, actively retain or remove information, or simply produce precise predictions at the right periods for time series data, including hydrological forecasts.

The following are the mathematical formulas that regulate the LSTM cell:

$$Ft = \sigma (Wf * [h(t-1), xt] + bf) \qquad (6)$$

$F_t$: Forget gate: It decides how much data from prior cell state should be kept or thrown away.

$$It = \sigma (Wi * [h(t-1), xt] + bi) \qquad (7)$$

$i_t$: input gate: Assess the cell state to see what additional data should be added from the current input ($x_t$).

$$Ot = \sigma (Wo * [h(t-1), xt] + bo) \qquad (8)$$

$O_t$: output gate: decides whether data from cell state should be sent to the $h_t$, the following concealed state.

$$Ct = ft * C(t-1) + it * tanh(Wc * [h(t-1), xt] + bc) \qquad (9)$$

$C_t$: Cell state: To update the memory cell, the candidate cell state is coupled with the output of the forget gate.

$$Ht = Ot * tanh(Ct) \qquad (10)$$

$H_t$: Hidden state

W: Weight matrices

b : bias vectors

σ: Sigmoid activation function

tanh: Hyperbolic tangent activation function

The data's long-term and short-term dependencies can be captured by LSTMs by combining these gates. Because they simulate sequential inputs like historical precipitation, discharge, and meteorological data to predict future water discharge, LSTMs are especially helpful in hydrological forecasting. LSTMs can better anticipate dynamic systems by modeling complicated relationships with the help of the non-linear activation functions (sigmoid and tanh).

- **Gated Recurrent Units (GRUs)**

GRUs are another type of recurrent neural network used to handle sequential data. They are also appropriate for timeseries forecasting tasks. However, the architecture of GRUs is simpler than LSTMs which makes it computationally efficient than LSTMs. The two gates, update and forget gate are used to control the information flow where the update gate tells the GRU how much of the previous hidden state to keep or release. While the reset gate informs to which extent, the model should disregard historical data. GRUs selectively recall or remove information while simulating complex temporal patterns through this process.

Although GRUs are simpler than LSTM, still these networks have demonstrated their efficiency in a range of time series forecasting tasks like hydrological forecasting. The GRU cell is determined by the following mathematical equations:

$$rt = \sigma(Wr * [h(t-1), x_t] + b_r br) \qquad (11)$$

$r_t$: Reset Gate: decides how much previous information should be ignored.

$$z_t = \sigma(W_z * [h_{(t-1)}, x_t] + b_z) \qquad (12)$$

$z_t$: Update Gate: Choose how much of the prior concealed state to keep or throw away.

$$h'_t = \tanh(W_h * [r_t * h_{(t-1)}, x_t] + b_h) \qquad (13)$$

h't: Candidate hidden state: candidate new state is based on the prior hidden state and the present input.

$$ht = (1 - z_t) * h_{(t-1)} + z_t * h't \qquad (14)$$

$h_t$: Hidden state: The final state of the GRU cell controlled by the update gate, is a function of the candidate hidden state and the prior hidden state.

GRUs has been able to simulate complex temporal patterns by selectively remembering or forgetting information by regulating the information flow through these gates, allowing it to produce precise predictions for time series data.

## 4. Experiment and Setup

A system equipped with an Intel Core i7 processor, a 64-bit operating system, and 4 GB of RAM for data preparation is used. Further, to build and train machine learning and deep learning time-series models Google Colab environment is used, using Python and TensorFlow packages with necessary libraries. It is used for both model designing and training and testing. The hyperparameter technique GridSearchCV is used to improve performance during model training. The test dataset is used to evaluate the performance of each trained model. To evaluate the performance of the model, assessment matrices are employed. The evaluation matrices used were Mean Absolute Error (MAE), Root Mean Square Error (RMSE), Mean Square Error (MSE), and R-square ($R^2$) calculated by Equations (15), (16),(17) and (18).

### 4.1. Mean Absolute Error (MAE)

MAE determines the average size of errors between the expected and actual values. It is helpful for datasets containing outliers since it is less susceptible to them than MSE and RMSE. It is calculated as follows:

$$MAE = 1/2 * \Sigma |yi - ŷi| \qquad (15)$$

### 4.2. Root Mean Square Error (RMSE)

It calculates the total squared difference between the predicted and actual values. It is sensitive to outliers as greater weight is assigned to larger errors such as outliers in RMSE. It is calculated as:

$$RMSE = sqrt((1/n) * \Sigma(yi - ŷi)^2) \qquad (16)$$

### 4.3. Mean Square Error (MSE)

MSE is used to estimate the average of the squared errors between actual and predicted values. To access the performance of regression and forecasting models it is a widely used matrix. It is extremely sensitive to outliers because of the squaring of the discrepancies, due to which it makes the larger errors more noticeable than the smaller ones. The formula of MSE is:

$$MSE = (1/n \, \Sigma(yi - ŷi)^2) \qquad (17)$$

### 4.4. Coefficient of Determination ($R^2$)

R2 score is a metric widely used in hydrological modeling. It is used to quantify how effectively a regression model can account for the variability of a dependent variable. It ranges from 0 to 1, with 1 denoting a robust forecast means that model is able to learn and predict the variability of dependent data very effectively and 0 denotes no explanation from the regression model indicating that the model is not able to learn for the variability of data. The formula for R2 is:

$$R^2 = 1 - SS_{tot}/SS_{res} \qquad (18)$$

where:

$y_i$: actual values,

$\hat{y}_i$: predicted values,

$\bar{y}$ is the mean of actual values,

$SS_{res} = \sum (y_i - \hat{y}_i)^2$

$SS_{tot} = \sum (y_i - \bar{y}_i)^2$

n : total number of data points.

## 5. Results and Discussion
### 5.1. Daily River flow Prediction Using Machine Learning Models Without Using River's Discharge feature as input

In this study, machine learning models SVR, XGBoost, and ANN were initially employed to predict River discharge using input features such as (Precipitation, Temperature Humidity, etc) without including discharge itself as a input variable. The SVR model obtained an RMSE of 454.50, MSE of 206,567.05, MAE of 293.65, and R-squared of 0.604, and the XGBoost model achieved an RMSE of 402.41, MSE of 161,933.53, MAE of 248.34, and R-squared of 0.690. The MLP model produced an RMSE of 580.85, MSE of 337,385.90, MAE of 417.83, and R-squared of 0.353 when discharge was not included in input features along with the climatic features. The GridSearchCV is used for the tuning of the model's hyper-parameter. The MLP with GridSearchCV tuning achieved RMSE of 367.12, MSE of 134,776.45, MAE of 230.92, and R-squared of 0.742 to predict discharge values still not using discharge as an input variable along with other features. All the models performed satisfactorily to forecast the daily river discharge by learning relationship between the climate variables and hydrological independent variables.

The results reveal that tuned MLP performed very well for regression problems as compared to other algorithms. It gains R2 of 0.74 indicating that it can account for the variability of the dependent variable effectively and it obtained low RMSE of 367.11 m3/sec compared to all other methods. Therefore, for the regression problem to learn the relationship between climate data and water data, ANN performed very well as shown in Table 1.

*Table 1: Comparison of Model For Daily River Discharge Prediction Without Using Discharge itself as Input*

| S.no | Model | MSE | RMSE | MAE | $R^2$ |
|---|---|---|---|---|---|
| 1 | SVR | 206567.051823 | 454.496482 | 293.651706 | 0.604050 |
| 2 | XGBOOST | 161933.531687 | 402.409657 | 248343675 | 0.689604 |
| 3 | ANN(MLP) | 337385.902672 | 580.849294 | 417.828928 | 0.353294 |
| 4 | MLP(GridSearchCV) | 134776.445724 | 367.119117 | 230.924415 | 0.741659 |

### 5.2. Daily Riverflow Prediction Using Machine Learning Models Using Lag Discharge Values

The lagged values of discharge from the previous five days were used with other independent climatic features in input to capture the temporal trend, to train the Machine learning models. The models take advantage of the data's temporal relationships by including input features with lagged values of water data, which is crucial for accurate time-series predictions. To optimize the XGBoost model's performance, a range of hyper-parameters were examined using GridSearchCV. Each configuration has been assessed by the grid search using the R2 scoring metric and 5-fold cross-validation. It examined several parameter combinations, including subsample ratios, column sampling ratios, learning rate, maximum tree depth, and the number of estimators. The model's optimal hyper-parameters, according to Grid-Search CV, were a maximum depth of 3, 100 estimators, 0.1 learning rate, 0.8 subsample ratio, and 1.0 column sampling ratio. XGBoost with tuned parameters achieved an RMSE of 184.05, an MAE of 79.19, an MSE of 33,875.86, and an R2 score of 0.9351.

The SVR model is used with hyper-parameter tuning through GridSearchCV, ensuring that 80 different hyper-parameter combinations were evaluated using 5-fold cross-validation. Whereas the regularization parameter C, kernel coefficient gamma, and the RBF kernel effective at modeling non-linear relationships were all set up to have a range of values in the parameter grid for SVR training. The optimal hyper-parameters found by GridSearchCV for SVR were kernel='rbf,' gamma=0.001, and C=100, to make predictions. After these parameters were adjusted, the SVR model achieved a Mean squared error (MSE) of 26,939.26, root mean squared error (RMSE) of 164.13, mean absolute error (MAE) of 71.32, and R-squared ($R^2$) value of 0.9484 on the test dataset. Then MLP is trained with the features based on factors such as lagged discharge values from the previous five days along other climatic variables. GridSearchCV was used to adjust the ANN model's architecture and hyper-parameters to determine the optimal number of hidden layer neurons, activation functions (ReLU and tanh), weight optimization solvers (Adam and lbfgs), learning rate, and the maximum number of iterations. To ensure a thorough analysis of every configuration, 5-fold cross-validation was utilized, and R2 was used as the scoring measure. The ideal configuration by Gridsearch for an ANN was a hidden layer with 50 neurons, ReLU activation function, Adam solver, constant learning rate, and up to 200 iterations. On the test set, the improved ANN model performed well with an R2 score of 0.9506, an MAE of 87.28, an RMSE of 160.57, and an MSE of 25,781.66. These results show how the optimized models using hyper-parameters tuning and the lagged input data could potentially be able to capture the temporal relationships between climate data and daily river flow.

Considering the lag variables of discharge for temporal trends along with the climate data improved the performance of the machine-learning model ensuring the discharge as a significant feature for temporal pattern learning. In this scenario, ANN outperforms all other algorithms having the highest R2 and lowest RMSE 160 shown in *Table 2*.

Table 2: Comparison of Models for Daily River Discharge Prediction Using Discharge itself as Input

| S.No | Model | MSE | RMSE | MAE | $R^2$ |
|---|---|---|---|---|---|
| 1 | SVR | 26939.255653 | 164.131824 | 71.322166 | 0.948371 |
| 2 | XGBOOST | 33875.864622 | 184.053972 | 79.186817 | 0.935077 |
| 3 | ANN(MLP) | 25781.656178 | 160.566672 | 87.283108 | 0.950589 |

### 5.3. Daily Riverflow Prediction through Deep Learning Models

The LSTM model is trained using a window of a 5-day data sequence of hydrological and meteorological variables to predict daily water discharge. By using this method, the model was able to identify patterns and temporal relationships in the data, which are essential for precise flood forecasting. A high correlation between the expected and actual values is demonstrated by the model's r-squared value of 0.9619, which shows that it explains 96.19% of the discharge variance. Moreover, the model's low prediction errors (RMSE of 140.96 and MAE of 73.45) demonstrate the model's reliability for accurate daily discharge prediction. Grid search was used for hyper-parameter optimization to maximize the model's performance. After experimenting with various batch sizes, epoch counts, and optimizers, the optimal setup was determined to be 16 epochs, batch sizes 10, and the 'sgd' optimizer. With a top cross-validation score of 0.9547, this combination demonstrated how well systematic adjustment may enhance the model's predictive ability. The LSTM successfully learned patterns from the input data and generated extremely accurate predictions because of a well-designed architecture and optimal hyperparameters.

Actual and anticipated discharge over time are compared in the graph shown in Figure 3. The LSTM model captures the general trend of the discharge data. The model is adept at the fundamental dynamics of the discharge process, as evidenced by the strong resemblance between the anticipated and real discharge, including peaks and troughs.

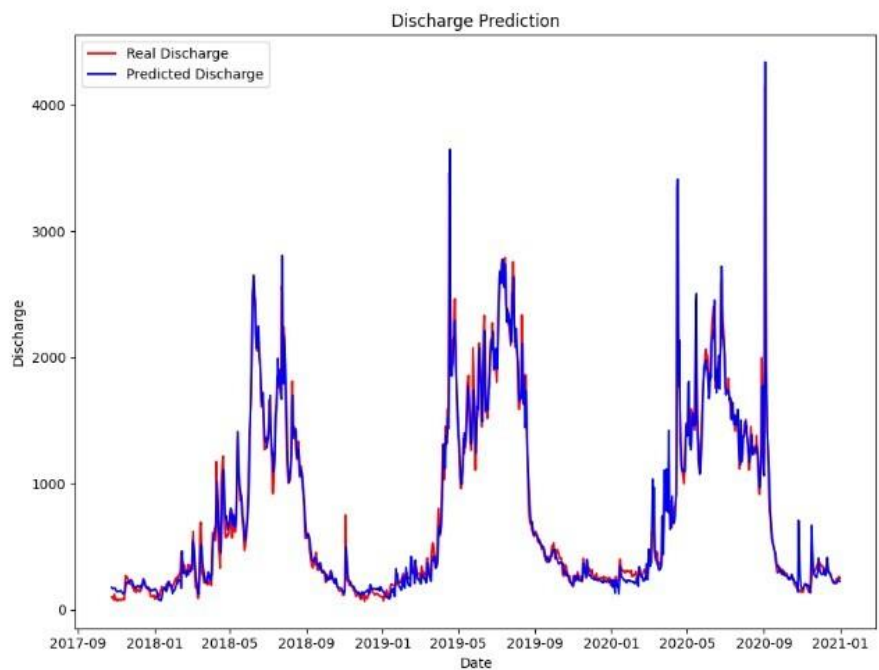

*Figure 3: Actual vs Daily Predicted Discharge through LSTM*

Although there are occasional variations, especially for larger discharge levels, the general pattern indicates that the LSTM model offers a trustworthy and accurate depiction of the discharge data.

Then the Gated Recurrent Unit model was trained and assessed using a 5-day window of data sequence of hydrological and meteorological variables to forecast daily water discharge. The model can explain around 95.51% of the variation in the discharge data, according to its R2 of 0.9551. This clearly suggests that it is a strong predictive model. There are moderate errors with an RMSE of 153.03 and MAE of 75.93, meaning that while the model was successful in identifying all the patterns, it was not quite as accurate as LSTM. To determine the optimum configuration for GRU performance, hyper-parameter optimization was carried out. A grid search has been performed using various optimizers, epochs, batch sizes, and parameter combinations. A batch size of 32, 20 epochs, and the 'Adam' optimizer are shown to be the optimal combination. The model obtained a cross-validation score of 0.9585 with this combination of hyper-parameters. The GRU model's capacity to represent the general dynamics of discharge is seen in Figure 4. The real discharge and the anticipated discharge nearly match, accurately capturing the sudden fluctuations and slow variations shown in the data. This indicates that the underlying temporal patterns in the discharge time series have been learned by the model.

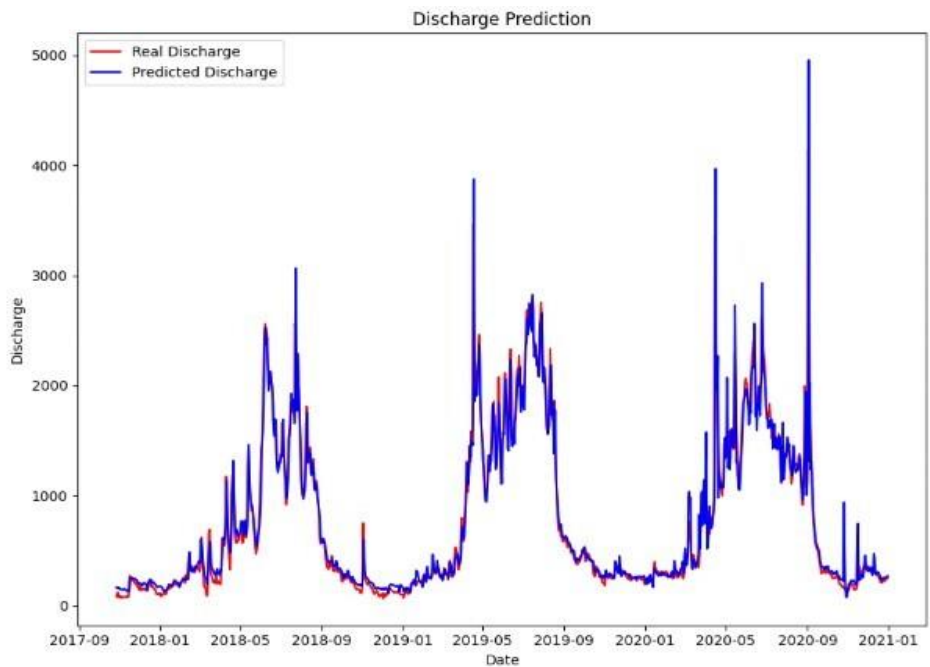

*Figure 4: Actual vs Daily Predicted Discharge through GRU*

With its capacity to analyze the temporal relationships present in sequential data, the GRU model generally performed well in predictions. In the present case, it was notably even somewhat less accurate than an LSTM. It is effective for flooding predictions due to its efficiency and similar degree of accuracy.

The results shown in **Error! Reference source not found.** indicate that the LSTM model performs best of the five models by reporting the lowest MSE, at 19870.19, RMSE, at 140.96; and MAE, at 73.45.

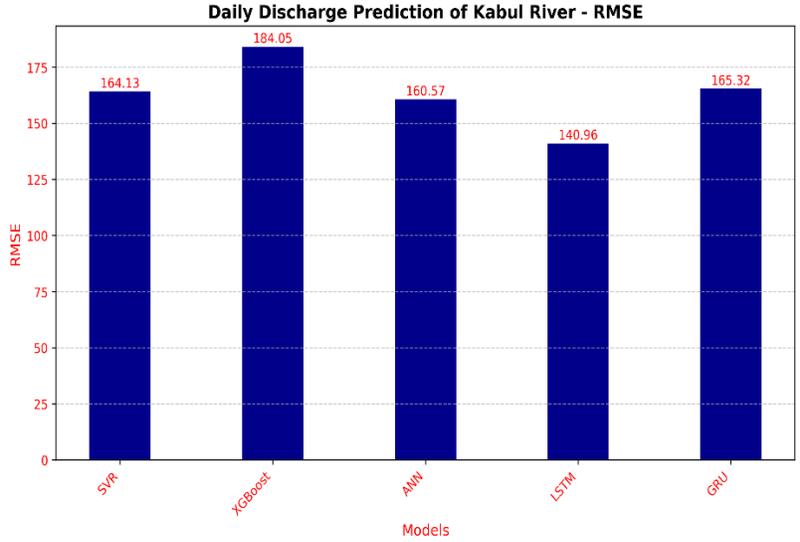

*Figure 5: Comparison of All Models in Terms of RMSE for Daily Discharge Prediction*

In addition, the highest R-squared, at 0.9619, implies that the LSTM model effectively captures the data's pattern and variance in data as shown in **Error! Reference source not found.**. The ANN model is the second best model with a little higher error and marginally lower R2, at 0.9506. Both the SVR and the GRU models have similar performance, with moderate errors and values around 0.948, less accurate than LSTM and ANN. XGBoost is the weakest, with the lowest R2 of 0.9351 and the highest errors. This can result from its inferiority to deep learning techniques in modeling sequential data. Overall, the LSTM is the best model for daily discharge prediction for the Kabul River.

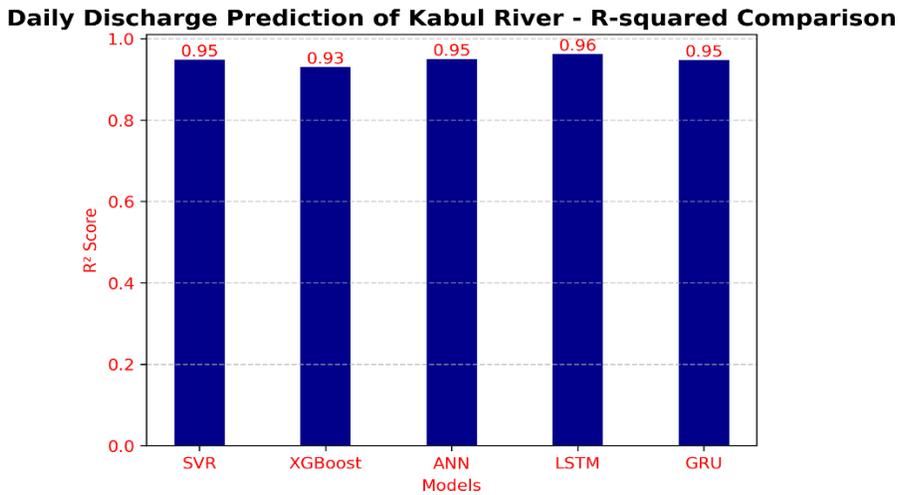

*Figure 6: Comparison of All Models in Terms of R2 for Daily Discharge Prediction*

Comparative results of all the models for daily discharge prediction using climate and water data are given in Table 3, which shows that LSTM outperformed all algorithms for daily Kabul River Discharge Prediction.

*Table 3: Comparison of All Models for Daily Discharge Prediction*

| S.No | Model | MSE | RMSE | MAE | $R^2$ |
|---|---|---|---|---|---|
| 1 | SVR | 26939.26 | 164.1318 | 71.32217 | 0.948371 |
| 2 | XGBOOST | 33875.86 | 184.054 | 79.18682 | 0.935077 |
| 3 | ANN | 25781.66 | 160.5667 | 87.283111 | 0.950589 |
| 5 | GRU | 27331.49 | 165.3224 | 76.60765 | 0.947652 |
| 4 | LSTM | 19870.19 | 140.9617 | 73.45287 | 0.961943 |

### 5.4. LSTM Model Performance Evaluation: Accurate Forecasting of Discharge Values

The models' performance evaluation shows that LSTM performs an outstanding task of forecasting the discharge levels. The model, according to its R-squared value of 0.96, can explain approximately 96% of the variation in the target variable. This illustrates how effectively the model is identifying the underlying patterns in the data. Additionally, the dataset's mean discharge of 865.48 m³/s and standard deviation of 775.68 m³/s, which represent around 19% of the mean discharge, make the Root Mean Square Error (RMSE) of 140 appear comparatively small. This accuracy level is in a reasonable range and demonstrates that the model is resilient to fluctuations in the data.

The MAE value of 73.5.00 m³/s indicates that, even though the discharge values of the dataset range from 64.8 m³/s to 9808.0 m³/s, and forecasts often deviate only marginally from actual values, further evidence the model's accuracy. These outcomes indicate the model's capacity to produce accurate forecasts despite the large range of discharge levels. Overall, the model's high R-squared, low RMSE, and MAE demonstrate that it works well and can accurately predict discharge levels in Kabul River KPK. The performance of LSTM in terms of errors and R2 is also compared with the previous model designed for river flow prediction in Kabul River, Pakistan. The comparison in Table 4 shows that LSTM outperformed the previous models for daily Kabul River discharge prediction in terms of RMSE and $R^2$.

*Table 4: Comparison of LSTM with other Models*

| S.No | Model | RMSE | $R^2$ |
|---|---|---|---|
| 1 | Hybrid(LASSO-ANN-EMD-SVM) | 333.1 | 0.789 |
| 2 | Random Forest Regressor | | 0.77 |
| 3 | Our LSTM Model | 140.9617 | 0.961943 |

### 5.5. Multistep Forecasting using Time Series Models

For multiple days ahead prediction sequential time series model LSTM and GRU models have trained and tested. As, sequential models are widely used and renowned in Deep learning fields for time series forecasting tasks.

The LSTM model is designed for 5-day multistep forecasting utilizing a window of 20 days of sequence data. The results displayed in Table 5 demonstrated a notable variance in performance throughout the prediction horizon. The model demonstrated dependable short-term predictions on Day 1 with the lowest MAE (1.3856), MSE (4.9850), and RMSE (2.2327), as well as the greatest R-squared (0.9378). However, by Day 5, performance has declined, with an MAE of 171.5339, an MSE of 86133.30, and an RMSE of 293.48, indicating the challenge of longer-term forecasting.

Even yet, the average R-squared value is still very high at 0.8438, suggesting that the model accounts for a significant portion of the variation explained by all of the predicted days. Batch_size = 32, epochs = 10, and optimizer = "adam" were the optimal settings determined by Grid Search CV, ensuring that the setup is optimized while balancing performance and training efficiency. Long-term forecasts might be improved with more modifications or improvements, including adding more features or sophisticated structures, as shown by the increasing inaccuracy with longer forecast horizons.

*Table 5:Multistep forecasting through LSTM*

| S.No | Day | MAE | MSE | RMSE | $R^2$ |
|---|---|---|---|---|---|
| 0 | 1 | 1.385599 | 4.984992 | 2.232710 | 0.937804 |
| 1 | 2 | 0.444717 | 0.602531 | 0.776229 | 0.912807 |
| 2 | 3 | 2.841203 | 26.026081 | 5.101576 | 0.882230 |
| 3 | 4 | 0.195466 | 0.114517 | 0.338403 | 0.861409 |
| 4 | 5 | 171.533897 | 86133.302984 | 293.484758 | 0.835768 |

The GRU is trained and tested for 5-day multistep forecasting with a window of 20-day sequence days, the model performed well in the short term but had a sharp rise in inaccuracy over longer time horizons. With an MAE of 0.9585, MSE of 3.5539, RMSE of 1.8852, and the highest R-squared (0.9556) on Day 1, the model demonstrated reliability and produced accurate short-term forecasts.

With the MAE at 0.3710 on Day 2, the performance remains considered excellent. By Day 3, the errors have significantly increased (MAE = 2.4828, RMSE = 4.8075), and on Day 5, the MAE and RMSE have increased to 152.7037 and 269.2061, respectively, which corresponds to an R-squared value of 0.8614 as shown in Table 6. The Grid Search CV indicated that batch_size=16, epochs=10, and optimizer 'Adam' were the best configurations. Such a setup is optimal and operating at peak efficiency. Even though the GRU model's performance might seem good for short-term horizons, the exponentially higher error over longer periods indicates that improvement could be achieved by using enhanced techniques to leverage advanced techniques for long-term predictability or by using a more refined sequence length in addition to highly relevant features.

*Table 6:Multi-step forecasting through GRU*

| S.no | Day | MAE | MSE | RMSE | $R^2$ |
|---|---|---|---|---|---|
| 0 | 1 | 0.958501 | 3.553855 | 1.885167 | 0.955578 |
| 1 | 2 | 0.371031 | 0.533373 | 0.730324 | 0.922644 |
| 2 | 3 | 2.482823 | 23.112192 | 4.807514 | 0.895151 |
| 3 | 4 | 0.170670 | 0.096565 | 0.310749 | 0.882804 |
| 4 | 5 | 152.703733 | 72471.944172 | 269.206137 | 0.861384 |

### 5.6. Comparison of Time Series Models for Multistep Forecasting

The Performance comparisons between LSTM and GRU models highlight their differences across a 5-day prediction horizon. Consistently high R2 values close to 1 demonstrate the excellent accuracy of the river discharge, although in most situations, GRU's R2 is marginally higher than LSTM's, indicating a superior capture of the data patterns. The two models' minimal RMSE error for Days 1–4 further demonstrates their reliability in making short-term forecasts. On Day 5, however, the RMSE for both models noticeably rises, and GRU performs somewhat better than LSTM. This pattern suggests that while both models are doing well, GRU performs more effectively, particularly in longer-term forecasts, when LSTM's performance is reduced.

### 6. Conclusions

Flood forecasting is crucial for reducing the destructive impact of river floods, which may have a catastrophic impact on infrastructure, economy, and people. Due to climate change and extreme rising temperature, sea levels are rising and unprecedented rains are occurring. Hence, flood prediction is a global Challenge. Therefore, reliable flood forecasting is required to provide early warnings and enable timely preparedness to reduce the impact of flooding. In the contemporary study, a case study is conducted on Kabul River to design a flood-forecasting model aimed at reducing the impact of floods and improving early warning systems. Therefore, the machine-learning and deep learning models are developed to predict river flow for the short term using satellite-based climate data and hydrological data. The AI-based flood prediction enables proactive disaster response efforts and enhances water resources management and early warning systems. Despite the complex nature of the region, the machine-learning models used in this study demonstrated significant performance in predicting river discharge. The outcomes reveal how machine-learning techniques can be used for flood prediction problems, especially in situations where data is scarce and conventional techniques are ineffective.

These models accurately predict daily river discharge and also for multiple days in advance. The model can be utilized as a reliable tool for anticipating flood events, making plans, and enhancing water management strategies by disaster management teams (NDMA). Acquiring precise and consistent data in transboundary rivers is a significant challenge in flood forecasting because of logistical and geopolitical issues. Despite these challenges, the research notably utilized satellite data to encapsulate the value of remote sensing technologies in flood prediction systems and produced significant results. These machine-learning models facilitate the minimization of the unpropitious effects of floods, preserving exposed populations and enhancing the resilience of flood-risk areas. The study highlights the significance of the integration of innovative data-driven techniques for flood prediction to create robust disaster management strategies. Although this study yields notable results, larger datasets will be helpful to improve future research for long-term forecasts. As well as upstream river discharge data, which provides the key insights of flood dynamics, it would also be to make accurate predictions for complex transboundary rivers, such as the Kabul River. The lack of upstream data for the discharge in this model prevented it from taking into consideration the impact of upstream changes on the flows and flooding at a location. It depicted the necessity of countries collaborating to exchange data regarding common basins. To build even more dependable and efficient flood forecasting systems, additional research may be conducted by recognizing these constraints and broadening the range of data sources.

**Dataset**

Climate data can be retrieved from the NASA Power website, while hydrological data is retrieved from WAPDA, which is not allowed to be shared publicly. Thus, it can be requested from WAPDA officials for research purposes.

**Acknowledgment**

We are also thankful to Mr. Abdul Hanan Tanveer, the representative of WAPDA, for his help in understanding the hydrological data. We would like to thank the Deputy Manager of Hydrology at NDMA for offering crucial insights into the study area.

**References**

*2022 Pakistan Floods*. (2023, September 6). Center for Disaster Philanthropy. https://disasterphilanthropy.org/disasters/2022-pakistan-floods/

Aatif, K., Fahiem, M. A., & Tahir, F. (2024). Forecasting Floods Using Deep Learning Models: A Longitudinal Case Study of Chenab River, Pakistan. *IEEE Access*. https://ieeexplore.ieee.org/abstract/document/10638656/

Ali, M., Taha, M., Aziz, M. S., Ahmed, H., & Ahmed, H. (2024). Flash Flood prediction of Panjkora River, KPK, Using Artificial Neural Networks (ANN) and Support Vector Machine (SVM). *Technical Journal*, *3*(ICACEE), 758–769.

Anuruddhika, M. L. P., Perera, K. K. K. R., Premarathna, L. P. N. D., Hansameenu, W. P. T., & Weerasinghe, V. P. A. (2025). A review of river flood models: Methods and applications for forecasting and simulation. *Ceylon Journal of Science*, *54*(1), 317–338. https://doi.org/10.4038/cjs.v54i1.8286

Atashi, V., Gorji, H. T., Shahabi, S. M., Kardan, R., & Lim, Y. H. (2022). Water level forecasting using deep learning time-series analysis: A case study of red river of the north. *Water*, *14*(12), 1971.

Atashi, V., Kardan, R., Gorji, H. T., & Lim, Y. H. (2023). Comparative Study of Deep Learning LSTM and 1D-CNN Models for Real-time Flood Prediction in Red River of the North, USA. *2023 IEEE International Conference on Electro Information Technology (eIT)*, 022–028. https://ieeexplore.ieee.org/abstract/document/10187358/

Aziz, A. (2014). Rainfall-runoff modeling of the trans-boundary Kabul River basin using Integrated Flood Analysis System (IFAS). *Pakistan Journal of Meteorology Vol*, *10*(20). https://www.pmd.gov.pk/rnd/rndweb/rnd_new/journal/vol10_issue20_files/7.pdf

Babar, M., Rani, M., & Ali, I. (2022). A Deep learning-based rainfall prediction for flood management. *2022 17th International Conference on Emerging Technologies (ICET)*, 196–199. https://ieeexplore.ieee.org/abstract/document/10004663/

Baig, S., & Hasson, S. ul. (2023). Flood Inundation and Streamflow Changes in the Kabul River Basin under Climate Change. *Sustainability*, *16*(1), 116.

Barez, E. (2023). *Determination of Flood Risk Areas and Development of Mitigation Strategies in Kabul River Basin, Afghanistan* [Master's Thesis, Izmir Institute of Technology (Turkey)]. https://search.proquest.com/openview/03e2b1f00e84b9f46345b8a22bb96815/1?pq-origsite=gscholar&cbl=2026366&diss=y

Bazarov, D., Ahmadi, M., Ghayur, A., & Vokhidov, O. (2023). The Kabul River Basin—The source of the Naglu and other reservoirs. *E3S Web of Conferences*, *365*, 03047. https://doi.org/10.1051/e3sconf/202336503047

Chen, J., Li, Y., Zhang, C., Tian, Y., & Guo, Z. (2023). Urban flooding prediction method based on the combination of LSTM neural network and numerical model. *International Journal of Environmental Research and Public Health*, *20*(2), 1043.

CRED. (2024). *Publications*. https://www.emdat.be/publications/

*DAWN*. (2024, September 10). DAWN.COM. https://www.dawn.com/news/1858033

Din, S. U. (2024). *Flow prediction in Kabul River: An artificial intelligence based technique*. https://www.allmultidisciplinaryjournal.com/uploads/archives/20240714130419_B-24-193.1.pdf

Dtissibe, F. Y., Ari, A. A. A., Abboubakar, H., Njoya, A. N., Mohamadou, A., & Thiare, O. (2024). A comparative study of Machine Learning and Deep Learning methods for flood forecasting in the Far-North region, Cameroon. *Scientific African*, *23*, e02053.

Dtissibe, F. Y., Ari, A. A. A., Titouna, C., Thiare, O., & Gueroui, A. M. (2020). Flood forecasting based on an artificial neural network scheme. *Natural Hazards*, *104*(2), 1211–1237. https://doi.org/10.1007/s11069-020-04211-5

El-Magd, S. A. A., Pradhan, B., & Alamri, A. (2021). Machine learning algorithm for flash flood prediction mapping in Wadi El-Laqeita and surroundings, Central Eastern Desert, Egypt. *Arabian Journal of Geosciences*, *14*(4), 323. https://doi.org/10.1007/s12517-021-06466-z

Fowler, K., Knoben, W., Peel, M., Peterson, T., Ryu, D., Saft, M., Seo, K., & Western, A. (2020). Many Commonly Used Rainfall-Runoff Models Lack Long, Slow Dynamics: Implications for Runoff Projections. *Water Resources Research*, *56*(5), e2019WR025286. https://doi.org/10.1029/2019WR025286

Gelaro, R., McCarty, W., Suárez, M. J., Todling, R., Molod, A., Takacs, L., Randles, C., Darmenov, A., Bosilovich, M. G., Reichle, R., Wargan, K., Coy, L., Cullather, R., Draper, C., Akella, S., Buchard, V., Conaty, A., da Silva, A., Gu, W., … Zhao, B. (2017). The Modern-Era Retrospective Analysis for Research and Applications, Version 2 (MERRA-2). *Journal of Climate*, *Volume 30*(Iss 13), 5419–5454. https://doi.org/10.1175/JCLI-D-16-0758.1

Gharakhanlou, N. M., & Perez, L. (2023). Flood susceptible prediction through the use of geospatial variables and machine learning methods. *Journal of Hydrology*, *617*, 129121.

Gunnam, G. R., Inupakutika, D., Mundlamuri, R., Kaghyan, S., & Akopian, D. (2023). Data-driven approach for robust flood prediction. *Electronic Imaging*, *35*, 1–5.

Halim, M. H., Wook, M., Hasbullah, N. A., Razali, N. A. M., & Hamid, H. E. C. (2022). Comparative assessment of data mining techniques for flash flood prediction. *Int J Adv Soft Comput Appl*, *14*. http://ijasca.zuj.edu.jo/PapersUploaded/2022.1.9.pdf

Hayder, I. M., Al-Amiedy, T. A., Ghaban, W., Saeed, F., Nasser, M., Al-Ali, G. A., & Younis, H. A. (2023). An intelligent early flood forecasting and prediction leveraging machine and deep learning algorithms with advanced alert system. *Processes*, *11*(2), 481.

Hussain, A., Cao, J., Ali, S., Ullah, W., Muhammad, S., Hussain, I., Rezaei, A., Hamal, K., Akhtar, M., Abbas, H., Wu, X., & Zhou, J. (2022). Variability in runoff and responses to land and oceanic parameters in the source region of the Indus River. *Ecological Indicators*, *140*, 109014. https://doi.org/10.1016/j.ecolind.2022.109014

Hussain, D., & Khan, A. A. (2020). Machine learning techniques for monthly river flow forecasting of Hunza River, Pakistan. *Earth Science Informatics*, *13*(3), 939–949. https://doi.org/10.1007/s12145-020-00450-z

Iqbal, M. S., Dahri, Z. H., Querner, E. P., Khan, A., & Hofstra, N. (2018). Impact of climate change on flood frequency and intensity in the Kabul River Basin. *Geosciences*, *8*(4), 114.

Islamic Relief. (2022). *Rapid Need Assessment Flood Emergency—Balochistan & Sind (28th Aug 2022)*. Islamic Relief. https://reliefweb.int/report/pakistan/rapid-need-assessment-flood-emergency-balochistan-sind-28th-aug-2022

John, T. J., & Nagaraj, R. (2023). Prediction of floods using improved pca with one-dimensional convolutional neural network. *International Journal of Intelligent Networks*, *4*, 122–129.

Kasyanto, H., Sari, R. R., & Lubis, M. F. (2023). Application of multilayer perceptron (MLP) method for streamflow forecasting (case study: Upper Citarum River, Indonesia). *IOP Conference Series: Earth and Environmental Science*, *1203*(1), 012032. https://iopscience.iop.org/article/10.1088/1755-1315/1203/1/012032/meta

Khaliq, S., Ali Khan, M., Bacha, U., Ali Taj, I., & Fazal. (n.d.). *Rescue operations underway across Pakistan as high-level floods in Indus, Kabul rivers wreak havoc*. DAWN.COM. Retrieved January 1, 2025, from https://www.dawn.com/news/1707042

Khan, S., Khan, M., Khan, A. U., Khan, F. A., Khan, S., & Fawad, M. (2023). Monthly streamflow forecasting for the Hunza River Basin using machine learning techniques. *Water Practice & Technology*, *18*(8), 1959–1969.

Luppichini, M., Vailati, G., Fontana, L., & Bini, M. (2024). Machine learning models for river flow forecasting in small catchments. *Scientific Reports*, *14*(1), 26740. https://doi.org/10.1038/s41598-024-78012-2

Musarat, M. A., Alaloul, W. S., Rabbani, M. B. A., Ali, M., Altaf, M., Fediuk, R., Vatin, N., Klyuev, S., Bukhari, H., & Sadiq, A. (2021). Kabul river flow prediction using automated ARIMA forecasting: A machine learning approach. *Sustainability*, *13*(19), 10720.

NASA. (2011, April 6). *Heavy Rains and Dry Lands Don't Mix: Reflections on the 2010 Pakistan Flood* [Text.Article]. NASA; NASA Earth Observatory. https://earthobservatory.nasa.gov/features/PakistanFloods

NASA,GMAO. (2024). *GMAO - Global Modeling and Assimilation Office Research Site*. NASA. https://gmao.gsfc.nasa.gov/

Nevo, S., Morin, E., Gerzi Rosenthal, A., Metzger, A., Barshai, C., Weitzner, D., Voloshin, D., Kratzert, F., Elidan, G., & Dror, G. (2022). Flood forecasting with machine learning models in an operational framework. *Hydrology and Earth System Sciences*, *26*(15), 4013–4032.

Nguyen, H. T., Duong, T. Q., Nguyen, L. D., Vo, T. Q., Tran, N. T., Dang, P. D., Nguyen, L. D., Dang, C. K., & Nguyen, L. K. (2020). Development of a spatial decision support system for real-time flood early warning in the Vu Gia-Thu Bon river basin, Quang Nam Province, Vietnam. *Sensors*, *20*(6), 1667.

*Pakistan: Floods | ReliefWeb*. (2022, July). ReliefWeb. https://reliefweb.int/disaster/fl-2022-000254-pak

Panahi, M., Jaafari, A., Shirzadi, A., Shahabi, H., Rahmati, O., Omidvar, E., Lee, S., & Bui, D. T. (2021). Deep learning neural networks for spatially explicit prediction of flash flood probability. *Geoscience Frontiers*, *12*(3), 101076.

Park, K., Jung, Y., Seong, Y., & Lee, S. (2022). Development of deep learning models to improve the accuracy of water levels time series prediction through multivariate hydrological data. *Water*, *14*(3), 469.

Rostami, A., & Gholizadeh, N. (2023). *Machine Learning and Deep Learning Approaches for River Flow Forecasting*. https://www.researchgate.net/profile/Amin-Rostami-8/publication/375722851_Machine_Learning_and_Deep_Learning_Approaches_for_River_Flow_Forecasting/links/655855acb86a1d521bf1f783/Machine-Learning-and-Deep-Learning-Approaches-for-River-Flow-Forecasting.pdf

Saeed, M., Li, H., Ullah, S., Rahman, A., Ali, A., Khan, R., Hassan, W., Munir, I., & Alam, S. (2021). Flood hazard zonation using an artificial neural network model: A case study of Kabul River Basin, Pakistan. *Sustainability*, *13*(24), 13953.

SAIS. (2017, August 22). *SAIS Review of International Affairs*. https://saisreview.sais.jhu.edu/water-crisis-in-kabul-could-be-severe-if-not-addressed/

Sayama, T., Ozawa, G., Kawakami, T., Nabesaka, S., & Fukami, K. (2012). Rainfall–runoff–inundation analysis of the 2010 Pakistan flood in the Kabul River basin. *Hydrological Sciences Journal*, *57*(2), 298–312. https://doi.org/10.1080/02626667.2011.644245

Shabbir, M., Chand, S., & Iqbal, F. (2023). A novel hybrid framework to model the relationship of daily river discharge with meteorological variables. *Meteorology Hydrology and Water Management. Research and Operational Applications*, *11*(2), 70–94.

Song, T., Ding, W., Wu, J., Liu, H., Zhou, H., & Chu, J. (2019). Flash flood forecasting based on long short-term memory networks. *Water*, *12*(1), 109.

*Statista*. (2024). Statista. https://www.statista.com/statistics/1306264/countries-most-exposed-to-floods-by-risk-index-score/

*UN Pakistan country report*. (2023). 37.

Ushiyama, T., Sayama, T., Tatebe, Y., Fujioka, S., & Fukami, K. (2014). Numerical simulation of 2010 Pakistan flood in the Kabul River basin by using lagged ensemble rainfall forecasting. *Journal of Hydrometeorology*, *15*(1), 193–211.

*WHO*. (2024). Floods. WHO. https://www.who.int/health-topics/floods

*WorldRiskReport*. (n.d.). IFHV. Retrieved January 1, 2025, from https://www.ifhv.de/publications/undefined

*WorldRiskReport*. (2024). IFHV. https://www.ifhv.de/publications/world-risk-report

Yaseen, A., Lu, J., & Chen, X. (2022). Flood susceptibility mapping in an arid region of Pakistan through ensemble machine learning model. *Stochastic Environmental Research and Risk Assessment*, *36*(10), 3041–3061. https://doi.org/10.1007/s00477-022-02179-1

Yaseen, M. W., Awais, M., Riaz, K., Rasheed, M. B., Waqar, M., & Rasheed, S. (2022). Artificial Intelligence Based Flood Forecasting for River Hunza at Danyor Station in Pakistan. *Archives of Hydro-Engineering and Environmental Mechanics*, *69*(1), 59–77. https://doi.org/10.2478/heem-2022-0005

Zhou, Q., Teng, S., Situ, Z., Liao, X., Feng, J., Chen, G., Zhang, J., & Lu, Z. (2023). A deep-learning-technique-based data-driven model for accurate and rapid flood predictions in temporal and spatial dimensions. *Hydrology and Earth System Sciences*, *27*(9), 1791–1808.